\documentclass[conference]{IEEEtran}
\IEEEoverridecommandlockouts
\usepackage[top=0.75in, bottom=1.1in, left=0.725in, right=0.725in]{geometry}

\ifCLASSINFOpdf
\else
 \fi
\usepackage{array}
\usepackage{hyperref}
\usepackage{lineno}
\modulolinenumbers[5]
\usepackage{amsmath,amssymb,amsfonts}
\usepackage{amsthm}

\usepackage{multirow}
\usepackage{graphicx}
\let\oldnl\nl% Store \nl in \oldnl
\newcommand{\nonl}{\renewcommand{\nl}{\let\nl\oldnl}}

\newtheorem{lemma}{Lemma}
\newtheorem{theorem}{Theorem}

\newtheorem{definition}{Definition}
\usepackage[noadjust]{cite}
\usepackage[font={footnotesize}]{caption}
\usepackage[font={footnotesize}]{subcaption}

\usepackage{algorithm}
\usepackage{algorithmic}
\usepackage{bm}
\usepackage{enumitem}
\usepackage{pifont}% http://ctan.org/pkg/pifont

\usepackage{xcolor}    % colors

\usepackage[]{threeparttable}
\usepackage{relsize}

\makeatletter
\newcommand\bigDiamond{\mathop{\mathpalette\bigDi@mond\relax}}
\newcommand\bigDi@mond[2]{%
  \vcenter{\hbox{\m@th
    \scalebox{\ifx#1\displaystyle 2\else1.2\fi}{$#1\Diamond$}%
  }}%
}
\newcommand\bigLozenge{\mathop{\mathpalette\bigL@zenge\relax}}
\newcommand\bigL@zenge[2]{%
  \vcenter{\hbox{\m@th
    \scalebox{\ifx#1\displaystyle 2\else1.2\fi}{$#1\blacklozenge$}%
  }}%
}
\makeatother

\begin{document}

\title{%
{\color{gray} \footnotesize
This paper has been accepted in the 21$^{st}$ Annual International Conference on Distributed Computing in Smart Systems and the Internet of Things, 2025} \\[1ex]
\textit{ReinDSplit}: Reinforced Dynamic Split Learning for Pest Recognition in Precision Agriculture
}

\author{
	\IEEEauthorblockN{Vishesh Kumar Tanwar\IEEEauthorrefmark{1}, Soumik Sarkar\IEEEauthorrefmark{2}, Asheesh K. Singh\IEEEauthorrefmark{3}, Sajal K. Das\IEEEauthorrefmark{1}}
	 \IEEEauthorblockA{
  \IEEEauthorrefmark{1}Department of Computer Science, Missouri University of Science and Technology, USA \\\IEEEauthorrefmark{2} Department of Mechanical Engineering, Iowa State University, USA\\ \IEEEauthorrefmark{3} Department of Agronomy, Iowa State University, USA\\
  vishesh.tanwar@mst.edu, soumiks@iastate.edu, singhak@iastate.edu, sdas@mst.edu\\
}}

\maketitle
\begin{abstract}
To empower precision agriculture through distributed machine learning (DML), split learning (SL) has emerged as a promising paradigm, partitioning deep neural networks (DNNs) between edge devices and servers to reduce computational burdens and preserve data privacy. However, conventional SL frameworks' \textit{one-split-fits-all} strategy is a critical limitation in agricultural ecosystems where edge insect monitoring devices exhibit vast heterogeneity in computational power, energy constraints, and connectivity. This leads to straggler bottlenecks, inefficient resource utilization, and compromised model performance. Bridging this gap, we introduce \textit{ReinDSplit}, a novel reinforcement learning (RL)-driven framework that dynamically tailors DNN split points for each device, optimizing efficiency without sacrificing accuracy. Specifically, a Q-learning agent acts as an adaptive \textit{orchestrator}, balancing workloads and latency thresholds across devices to mitigate computational starvation or overload. By framing split layer selection as a finite-state Markov decision process, \textit{ReinDSplit} convergence ensures that highly constrained devices contribute meaningfully to model training over time. Evaluated on three insect classification datasets using ResNet18, GoogleNet, and MobileNetV2, \textit{ReinDSplit} achieves 94.31\% accuracy with MobileNetV2. Beyond agriculture, \textit{ReinDSplit} pioneers a paradigm shift in SL by harmonizing RL for resource efficiency, privacy, and scalability in heterogeneous environments.
\end{abstract}
\section{Introduction}\label{sec:intro}
By 2050, global agricultural production must double to feed 10 billion people—a Herculean task exacerbated by climate-induced disruptions to pest populations. Rising temperatures accelerate insect reproduction cycles, amplifying crop destruction: the UN Food and Agriculture Organization estimates annual losses of 20–40\% (\$70 billion) due to pests. These losses are further compounded by inefficient pest treatment strategies, often disregarding Integrated Pest Management (IPM) recommendations. While cutting-edge solar-powered insect monitoring tools have modernized entomological studies and pest control efforts, many rural agricultural edge devices operate with limited computational resources. They are unavailable due to poor network connectivity, delaying near real-time data analysis. This leads to pesticide overuse, increasing growers' input costs and jeopardizing ecological integrity through soil degradation, water contamination, and killing pollinators. With 150+ AgriTech interviews, the authors observed the need for resource-efficient and privacy-centric technologies, adaptability for heterogeneous devices, and delaying actionable insights by days~\cite{I-Corps2023,singh2024smart}.
\begin{figure}[!t]
		\begin{minipage}{\columnwidth}
		\centering			
\includegraphics[width=1.0\linewidth, height=9cm]{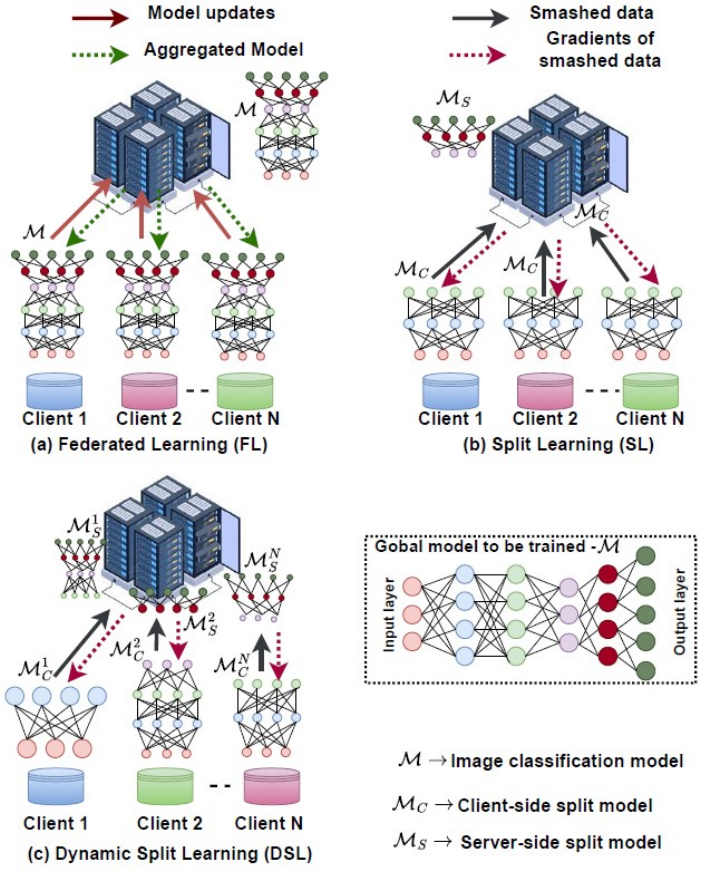}	
  % \vspace{-0.75cm}
		\caption{\small{Comparing DML scenarios of (a) FL, (b) SL, and (c) DSL (\textit{ReinDSplit}). In DSL, Client 2 may have more computational resources and client $N$ may have a longer time window.}}
  % \vspace{-0.5cm}
		\label{compare_fl_sl_dsl} 
		\end{minipage}
        \vspace{-0.275in}
\end{figure}

Federated Learning (FL)~\cite{antunes2022federated} emerged as a distributed machine learning (ML) by exchanging model gradients instead of raw data (Fig.~\ref{compare_fl_sl_dsl} (a)). However, FL mandates full DNN model training over the local edge devices, straining devices with limited compute or intermittent connectivity—a mismatch for agricultural edge nodes governed by solar cycles and sub-1 Mbps bandwidth~\cite{imteaj2021survey}, leading to inefficiencies, infeasibility, and a lower technology adoption rate. Though FL restricts the flow of raw data, gradient inversion attacks~\cite{mothukuri2021survey} can reconstruct sensitive farm data, eroding trust in FL's privacy guarantees. To address FL's limitations, Split Learning (SL)~\cite{vepakomma2018split} is a lightweight alternative for leveraging client data that cannot be centralized due to bandwidth limitations, high communication and computational overheads, and privacy concerns. In SL, a DNN model, denoted as $\mathcal{M}$, is split into two segments: $\mathcal{M}_C$ for the client's device and $\mathcal{M}_S$ for the server, as presented in Fig.~\ref{compare_fl_sl_dsl} (b). The client-side split, $\mathcal{M}_C$, contains the input layer upto the split point (a.k.a. cut layer), while the server-side split $\mathcal{M}_S$ comprises the remaining layers. Given $N$ participating clients in the SL framework, each client conducts a forward propagation (e.g., feature extractors) on their local data $\mathcal{M}_C$ upto the cut layer and sends intermediate activations (smashed data) to the server. Upon receiving the smashed data, the server completes the forward propagation and loss computation and commences backpropagation until the cut layer. Once the gradients are calculated, $\mathcal{M}_S$ send them to the respective client to complete the backpropagation and update the model parameters. This completes one round, and the cycle persists until convergence. 

While SL reduces client-side computing by 30–60\% than FL~\cite{bakhtiarnia2023dynamic}, its rigid \textit{same-split-for-all} design inherently assumes homogeneous device capabilities—a flawed premise in agriculture. Powerful drones idle with shallow splits, while farm-edge insect monitoring devices choke on deep ones, creating stragglers that delay global convergence. Static splits also ignore transient factors like battery drain or network congestion. These tailored challenges lead to \textit{powerful devices remaining underutilized, while weaker devices risk becoming performance bottlenecks, and the close coupling between the server and each device can exacerbate system fragility.} Recent work~\cite{cao2024learning} adapted splits per device; however, they rely on heuristic policies that lack theoretical guarantees and fail to scale with dynamic agriculture conditions. Given the increasing complexity of deep neural network (DNN) architectures, deterministically selecting an optimal, personalized cut layer (DNN-agnostic) for each resource-constrained edge device is challenging. These challenges are magnified when dynamic cut-layer identification and assignment are required across adaptive agricultural devices.
Motivated by reinforcement learning's (RL) ability to adapt to dynamic, uncertain environments via reward-driven exploration, our approach leverages RL to determine optimal cut-layer placements autonomously.

We employ a Q-learning agent that frames split selection as a finite Markov Decision Process (MDP), balancing computational load, latency, and accuracy. This adaptive framework overcomes agriculture-specific limitations, aligning with UN sustainable development goals and inherently protecting from adversarial model inversion attacks~\cite{mothukuri2021survey}, offering a blueprint for dynamic SL irrespective of underlined DNN (as shown in Fig.~\ref{compare_fl_sl_dsl} (c)) in applications such as smart healthcare and Industry 4.0. Overall, our solution effectively addresses device heterogeneity and dynamic resource management challenges. 
Our contributions are:
% \vspace{-0.2cm}
\begin{enumerate}%[leftmargin=12pt]
% \vspace{-0.15cm}
   \item Our proposed RL-based dynamic SL framework, abbreviated as \textit{ReinDSplit}, addresses the inherent limitations of the conventional SL, specifically accommodating heterogeneous devices. To our knowledge, \textit{ReinDSplit} is the first step towards adaptive SL using RL.%\vspace{-0.3cm}
    \item We provided theoretical analysis demonstrating that, with Q-learning on a finite MDP, the probability of choosing an infeasible split approaches zero, ensuring stable local gradients, straggler-free convergence, and improved resource utilization of heterogeneous devices.
    \item We conducted comprehensive experiments on \textit{ReinDSplit} usability (accuracy) and split-point assignment trade-off with model performance and client load for three pest recognition datasets with three SOTA DNN architectures: ResNet18, GoogleNet, and MobileNetV2.
    \item \textit{ReinDSplit} outperforms traditional SL and is comparable to FL by delivering superior computational efficiency and model accuracy (71–94\%) across IID and non-IID settings. MobileNetV2 achieves 94.31\% (IID), demonstrating robust tolerance to heterogeneous distributions. 
\end{enumerate}

\noindent \textbf{Organization:} Related work in Section~\ref{related_work} and preliminaries in Section~\ref{prelims}. Our system model and proposed methodology are discussed in Section~\ref{proposed_approach}, and theoretical analysis and experimental results in Section~\ref{theoretical_ana} and Section~\ref{experiments}, respectively. Section~\ref{conclusion} concludes the paper with future work.\vspace{-0.1cm}

\section{Related Work}\label{related_work}\vspace{-0.1cm}
\subsection{Pest Monitoring and Precision Agriculture}\vspace{-0.1cm}
Advances in sensing technologies and ML are driving precision pest monitoring and IPM strategies. Sensor- and image-based detection has demonstrated upto 97\% accuracies~\cite{lima2020automatic}. Further refinement is illustrated by~\cite{butera2021precise}, where a Faster R-CNN model with a MobileNetV3 backbone achieved an average precision of 92.66\% for insect detection under non-homogeneous conditions. Optical remote sensing methods, including UAV and satellite platforms~\cite{sishodia2020applications}, expand the spatial scope for real-time pest or disease surveillance. In particular,~\cite{kirkeby2021advances} collected 10K field records using optical sensors in oilseed rape crops, surpassing 80\% accuracy in classifying flying pests. UAVs are increasingly crucial in precision spraying and localized interventions~\cite{iost2020drones}. Beyond high-resolution imaging capabilities, they can be tailored for different agronomic tasks, as shown by~\cite{velusamy2021unmanned}, which investigated fixed-wing, single-rotor, and multi-rotor UAVs for targeted pest management. Real-time DNN frameworks have also emerged; ~\cite{albanese2021automated} offers continuous orchard surveillance with minimal power demands, though bandwidth constraints persist in large farms~\cite{chen2020aiot}. The scope of detection tasks extends beyond pests, with~\cite{anand2021agrisegnet} demonstrating anomaly segmentation in UAV images for weeds and other farmland disruptions. 
\vspace{-0.2cm}

\subsection{Split Learning}\vspace{-0.1cm}
FL and SL have gained attention to safeguard data privacy and reduce computational burdens in agriculture-focused IoT scenarios. Leveraging SL’s capacity for partial computation offloading,~\cite{pham2023binarizing} presents binarized local layers, cutting memory overhead and curtailing exposure of sensitive model intermediates with minimal performance decline. Further enhancements in distributed learning involve hybrid architectures. PPSFL~\cite{zheng2024ppsfl} merges FL and SL with private group normalization, effectively handling data heterogeneity and resisting gradient inversion attempts. Likewise, a federated split learning method in~\cite{zhang2023privacy} lowers the computational burden for client devices, retaining comparable accuracy to standard SL. Parallel SL designs have also been introduced for communication efficiency, as discussed in~\cite{lin2024efficient}, where channel allocation and gradient aggregation significantly cut overall latency. Heterogeneous client scenarios motivate ring-based SL strategies, with~\cite{shen2023ringsfl} mitigating slower “straggler” clients. \vspace{-0.15cm}

\subsection{RL-based Resource Allocation Strategies}\vspace{-0.1cm}
The interplay between device heterogeneity, fluctuating connectivity, and limited energy budgets in IoT networks necessitates robust resource allocation strategies. RL has emerged as a powerful tool for dynamic optimization, offering policy-driven adaptations at runtime. Concurrent Federated RL~\cite{tianqing2021resource} exemplifies this, merging FL principles with RL agents to improve system-wide utility and expedite task completions in edge computing. The approach outperforms classical baselines by jointly addressing privacy preservation and rapid decision-making. Similarly,~\cite{xiong2020resource} adopts a deep RL framework for mobile edge computing, reporting 15\% lower task completion times and a 20\% reduction in resource demands compared to standard DQN approaches.

Clustering-based offloading further refines performance, as demonstrated by~\cite{liu2020resource}, which outperforms system cost outcomes via an RL-driven grouping of IoT users. Additional complexities arise when handling general task graphs, prompting the advanced DRL scheme of~\cite{yan2020offloading} to reduce energy-time costs by upto 99.1\% of the theoretical optimum. DeepEdge~\cite{alqerm2021deepedge} similarly harnesses a two-stage RL scheme to improve QoE, enhancing latency and success rates for edge-based IoT workloads. A multi-agent perspective is highlighted in~\cite{liu2020multi}, where IL-based Q-learning yields a 25\% improvement in system costs by enabling distributed decision-making among selfish clients. Although these studies illustrate RL’s efficacy, concerns over high-dimensional state spaces and scalability persist in multi-farm or large-scale settings.\vspace{-0.15cm}

\section{Reinforcement Learning Overview}\label{prelims}
\subsubsection{Preliminaries}\label{prelim:rl}
A Markov Decision Process (MDP) is characterized by a tuple \((S, A, P, R, T)\), where $S$ denotes set of all possible states, $A$ denotes the finite set of all possible actions, $P: S \times A \to P(S)$ is the state transition probability function; $P(s_{t+1}\mid s_t, a_t)$ giving the probability of moving to state $s_{t+1}$ from state $s_t$ after acting $a_t$, $R: S \times A \times S \to \mathbb{R}$ is the reward function, where $R_t = R(s_t, a_t, s_{t+1})$ is the reward received by the agent upon transitioning from $s_t$ to $s_{t+1}$ via $a_t$, and $T$ is the terminal time. We seek a policy \(\pi_{\theta}\) (parametrized by \(\theta\)) that maximizes the cumulative, possibly discounted, future rewards in this MDP. Formally, if \(\gamma \in [0,1]\) is the discount factor, the expected return starting from state \(s_t\) and action \(a_t\) under policy \(\pi_{\theta}\) is captured by the state-action value function \(Q(s_t, a_t)\), defined as
\vspace{-0.3cm}
\begin{equation}
\vspace{-0.15cm}
Q(s_t, a_t) 
\;=\;
\mathbb{E}\!\Bigl[
\,\sum_{i=0}^{\infty}\gamma^i\,R_{t+1 + i}
% \;\middle|\;
\hspace{0.1cm}
s_t,\,a_t
\Bigr]
\label{eq:qvalue}
\vspace{-0.05cm}
\end{equation}
We aim to identify an optimal policy \(\pi_{\theta}^*\) that yields the highest \(Q\)-values possible, say $\pi_{\theta}^*
\;=\;
\arg\max_{\theta}\,Q(s_t,\,a_t)$.

In RL, an agent learns purely from trial-and-error experience, guided by reward signals rather than labeled examples. At each step, the agent in state \(s_t\) takes an action \(a_t \in A\). Upon executing \(a_t\), it transitions to a new state \(s_{t+1}\) and obtains a scalar reward \(R(s_t, a_t, s_{t+1})\). A \textit{policy} \(\pi\) maps each state to an action (or probability distribution over actions). The agent's main objective is to find an \textit{optimal policy} \(\pi^*\) that yields the maximum expected discounted return. Mathematically, 
\vspace{-0.15cm}
\begin{equation}
\pi^*(s) = \arg\max_{a \in A}
\,\gamma \sum_{s' \in S} P(s' \mid s, a)\,
V^*(s'),    
\vspace{-0.1cm}
\end{equation}
where \(V^*(s)\) is the optimal value function at state \(s\), satisfying
\vspace{-0.2cm}
\begin{equation}
V^*(s) = \max_{a \in A}
\,\Bigl(R(s, a) + \gamma\sum_{s' \in S} P(s' \mid s, a)\,V^*(s')\Bigr)    
\vspace{-0.2cm}
\end{equation}

These value functions capture how ``good'' it is to be in a particular state and help evaluate different policies.\vspace{0.1cm}

\subsubsection{Deep RL via Q-Learning}
An approximate function is used to represent \(Q\), especially for high-dimensional state or action spaces. We approach includes \textit{Deep Q-Network (DQN)}, which replaces the tabular Q-value storage with a neural network $Q_{\theta}(s, a)$. In DQN, our network takes as input the current state, $s_t$, and the output layer provides an estimate $Q_{\theta}(s_t, a)$ for each $a\in A$.\\\vspace{-0.25cm}

\noindent \textit{Loss Function:} At each time step \(t\), we observe a transition, \((s_t, a_t, R_{t+1}, s_{t+1})\) and the target for Q-learning is given by
\vspace{-0.15cm}
\begin{equation}
\vspace{-0.1cm}
y_t=R_{t+1} \;+\;\gamma\,\max_{\,a^\prime}\,Q_{\theta^{-}}(s_{t+1},\,a^\prime),
\label{eq:target}
\vspace{-0.1cm}
\end{equation}
where \(\theta^{-}\) represents the parameter set of a \textit{target network}, which is periodically updated (and remains fixed between updates to stabilize training). We calculate the DQN loss as \vspace{-0.1cm}
\begin{equation}
\vspace{-0.1cm}
L(\theta) = \Bigl(\,y_t \;-\; Q_{\theta}(s_t,\,a_t)
\Bigr)^{2},
\label{eq:dqnloss}
\end{equation}
where \(y_t\) is the target from \eqref{eq:target}, and \(Q_{\theta}(s_t,a_t)\) is the predicted Q-value from the DQN. The parameters \(\theta\) are updated by minimizing \(\sum_{t}L(\theta)\).% using gradient-based methods (e.g., stochastic gradient descent).
\vspace{-0.1cm}
\subsection{Why DQNs for Our Framework?}\vspace{-0.1cm}
Our framework operates in a discrete action space where each action designates a model cut layer under dynamic resource and time constraints. Because the state space-encompassing device resource availability, time windows, and partial model outputs are large and complex, tabular Q-learning becomes impractical. Instead, DQNs leverage neural function approximators to estimate Q-values within this complex space. By framing split selection as a finite MDP, our approach exploits RL’s reward-driven exploration to adapt to uncertain environments. Leveraging replay, a target network, and an $\epsilon$-greedy strategy, we balance exploration and exploitation, optimizing split assignments across devices.

\section{Our Proposed Framework}\label{proposed_approach}
This section discusses our system model, mathematical formulation, and proposed framework, \textit{ReinDSplit}. \vspace{-0.2cm}
\subsection{System Model}
\begin{figure}[!t]
		\begin{minipage}{\columnwidth}
		\centering			
  \includegraphics[width=1.0\linewidth]{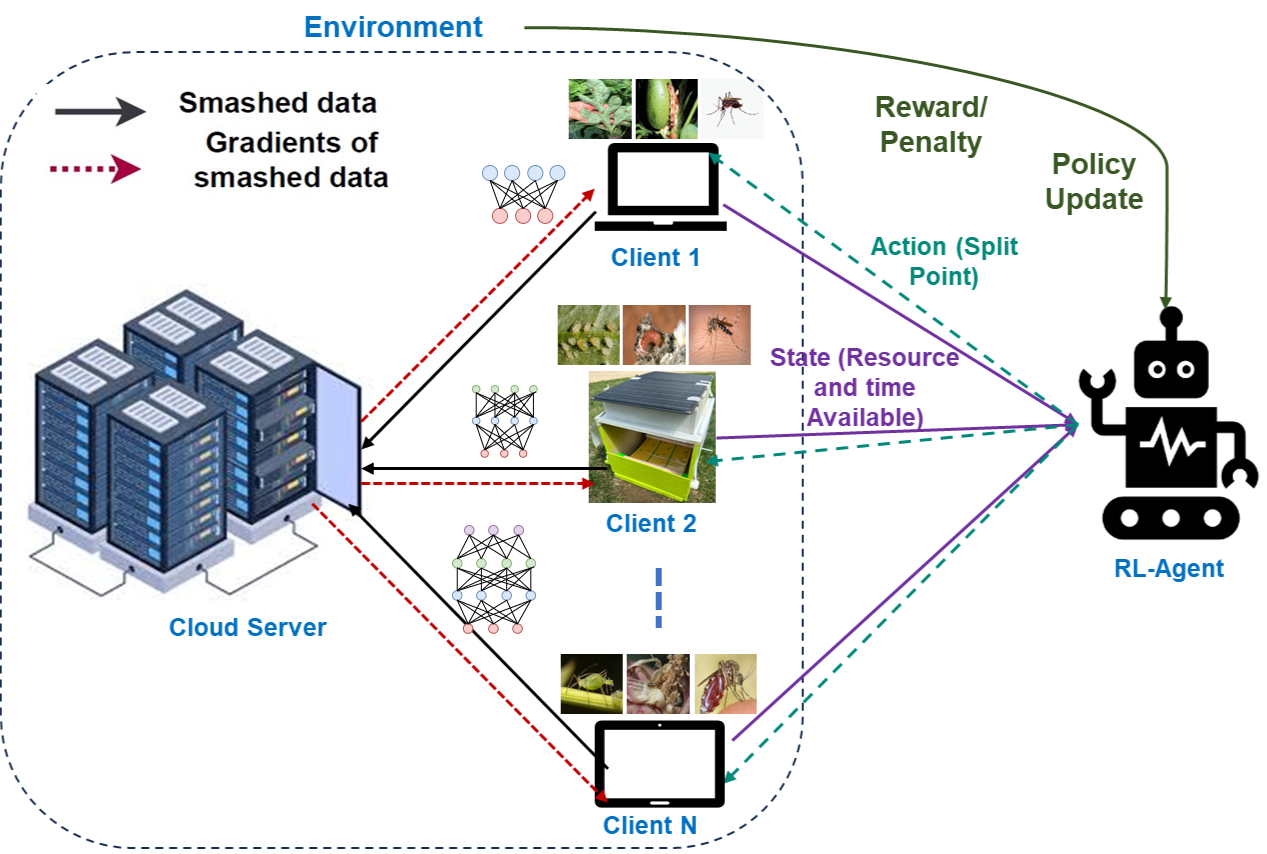}	
  % \vspace{-0.75cm}
		\caption{Overview of our proposed \textit{ReinDSplit}-based pest recognition system. Client 2's device is our in-lab-developed automated insect monitoring prototype and will be deployed for real-time field validation in our future work.}
  % \vspace{-0.5cm}
		\label{overview} 
		\end{minipage}
        \vspace{-0.3in}
\end{figure}

We consider an agricultural region $\mathcal{R}$, comprises $N$ geographically separated farms. In each location, a client or \textit{edge} device $d_i \in \mathcal{H} = \{d_1, d_2, \ldots, d_N\}$ captures high-resolution images of insect pests, as presented in Fig.~\ref{overview}. Though spatially apart, these farms cultivate the same crops (like soybeans or corn) under analogous conditions, thus exhibiting a near-\textit{IID} distribution of pest species. Each $d_i$ has limited computational resources $R_i$ (e.g., CPU/Jetson Nano) and a varying active time window $T_i$ due to solar battery life and power schedules. We assume an intermittent communication network exists (periodically slow and unreliable) for smashed data exchange. Contextually, we use device and client interchangeably.

A \textit{cloud server} $\mathcal{S}$ manages the SL framework, where a global DNN model $\mathcal{M}$ is partitioned into $K$ ``\textit{client-server}" submodel pairs denoted as:
\vspace{-0.1cm}
\begin{equation*}
\vspace{-0.1cm}
\Gamma =\Bigl\{\bigl(\mathcal{M}_C^1,\mathcal{M}_S^1\bigr),\,\bigl(\mathcal{M}_C^2,\mathcal{M}_S^2\bigr), \ldots, \bigl(\mathcal{M}_C^K,\mathcal{M}_S^K\bigr)\Bigr\},
\end{equation*}
where each $\mathcal{M}_C^k$ is computed locally on $d_i$ and the complementary part $\mathcal{M}_S^k$ executes on $S$. The server is aware and can estimate each submodel’s minimum requirements $R_{\text{required}}(\mathcal{M}_C^k)$ and $T_{\text{required}}(\mathcal{M}_C^k)$, along with local farm details, such as $R_i$, $T_i$ and dataset. By selecting an appropriate split point (or cut layer) $k$ for each $d_i$, the system aims to balance classification accuracy with heterogeneous computational and time constraints. If $\mathcal{M}_C^k$ demands more time than expected $T_i$, the device becomes a \textit{straggler}, leading to incomplete local training. Alternatively, our proposed framework aims to maximize aggregated accuracy as:
\vspace{-0.1cm}
\begin{equation}
\vspace{-0.1cm}
\label{eq:main_obj}
\begin{aligned}
\max_{\varphi}\quad 
& \sum_{i=1}^N \text{Acc}\bigl(\varphi(i)\bigr) \\
\text{subject to} \quad 
& R_{\text{required}}\bigl(\mathcal{M}_C^{\varphi(i)}\bigr)\,\le\,R_i, 
\quad \forall i = 1,\dots,N, \\
& T_{\text{required}}\bigl(\mathcal{M}_C^{\varphi(i)}\bigr)\,\le\,T_i, 
\quad \forall i = 1,\dots,N, \\
& \varphi(i)\in \{1,\dots,K\}, 
\quad \forall i = 1,\dots,N.
\end{aligned}
\end{equation}

where, $\varphi(i) = k$ is an \textit{allocation function} that decides the split model-pair $\bigl(\mathcal{M}_C^k,\mathcal{M}_S^k\bigr)$ to be assigned to the farm $d_i$ and $\text{Acc}\bigl(\varphi(i)\bigr)$ denotes the expected pest classification accuracy. In this paper, we constitute only two constraints; however, in our future work, we will formulate additional constraints such as battery budget, memory, bandwidth, etc., and objectives (e.g., minimizing average training time, balancing device load, or weighting accuracy per device).\vspace{-0.1cm}

\subsection{Reinforcement-based Dynamic SL (ReinDSplit)}\label{split_reinforcement_learning}

\textit{ReinDSplit} develops an adaptive policy for partial model allocation that considers dynamic resource and time constraints, precluding raw data sharing. Each client $d_i\in \mathcal{H}$  is viewed as an agent in an MDP, where $t^{th}$ state $s_t$ encodes (i) local resources $R_i$, (ii) time availability $T_i$, and (iii) partial model parameters for the client-side split $\mathcal{M}_C$. An action $a_t$ determines the current \textit{split point}. The reward $R_{t+1}$ derives from a performance objective (Eq.~\ref{eq:main_obj}) once the partial forward pass and backward propagation complete with $\mathcal{M}_C$ and $\mathcal{M}_S$.

By learning a Q-function mapping states to split-layer actions, \textit{ReinDSplit} adaptively selects $\mathcal{M}_C^k,\mathcal{M}_S^k$ for each round. This balances \textit{efficiency} (lightweight partial forward passes on constrained devices) and \textit{performance} (server-side layers benefit from aggregated gradient signals). Moreover, limited gradient transmission to the server and raw data always remaining local increases protection from adversarial attacks such as data reconstruction, thus enhancing privacy~\cite{mothukuri2021survey}. This unifies RLs' multi-agent perspective with adaptive model partitioning of SL to orchestrate \textit{ReinDSplit} in resource- and privacy-critical scenarios.

\subsection{Mathematical Formulation}\label{sec:math_formulation}\vspace{-0.1cm}
Let the overall training be divided into discrete rounds \(t = 1, 2, \dots\). We aim to learn a policy that adaptively assigns the split index \(\varphi(i)\) with constraints $R_i$ and $T_i$ (ref. Eq.~\eqref{eq:main_obj}).

\subsubsection{State Space}
At round $t$, the state of the device \(d_i\) is:
\vspace{-0.2cm}
\begin{equation}
\vspace{-0.2cm}
s_i^t \;=\;\bigl(R_i^t,\,T_i^t,\,\mathcal{P}_i^t\bigr),
\label{eq:state_space}
\end{equation}
\noindent where $R_i^t$ and $T_i^t$ represent the available computational resources and time window, respectively, for device \(d_i\), dedicated to local training in round \(t\), and $\mathcal{P}_i^t$ encapsulates partial model performance metric % or relevant performance metrics 
for the client-side model.

After each round, the environment (\textit{ReinDSplit} plus the device’s local conditions) transitions \(s_i^t \to s_i^{t+1}\) based on the chosen action \(a_i^t\) and the resource--time consumption of training, as highlighted in Fig.~\ref{overview}.

\subsubsection{Action Space}
At each round \(t\), device \(d_i\) selects an action \(a_i^t\) from the finite set $\mathcal{A}_i = \bigl\{\,1,\,2,\,\dots,\,K\bigr\},$ where each integer \(k \in \{1,\dots,K\}\) indicates choosing the split pair \(\bigl(\mathcal{M}_C^k,\,\mathcal{M}_S^k\bigr)\) for local processing. Here, \(a_i^t\!=\!k\) determines how many layers are executed on the client versus the server, thus dictating the local resource--time burden. 

\subsubsection{Reward Function}\label{subsubsec:reward}
Upon taking action \(a_i^t\) in state \(s_i^t\), the agent receives an immediate reward \(r_i^t\). We design \(r_i^t\) to balance classification performance and resource--time feasibility, aligning with the objective formulated in \eqref{eq:main_obj}:

\begin{equation}
% \vspace{-0.15cm}
\small
\label{eq:reward_function}
r_i^t =
\begin{cases}
\displaystyle
\alpha\,\text{Acc}\bigl(\varphi(i)\bigr)
-\beta\,
\Bigl(
\max\{0,\;R_{\text{required}}(\mathcal{M}_C^{\varphi(i)})-R_i^t\} \\
\hspace{20pt}+\max\{0,\;T_{\text{required}}(\mathcal{M}_C^{\varphi(i)})-T_i^t\}
\Bigr), 
& \hspace{-0.75cm} \text{if feasible}, \\[2pt]
\displaystyle
-\gamma\,(\text{penalty}), & \hspace{-0.75cm} \text{otherwise}
\end{cases}
% \vspace{-0.15cm}
\end{equation}

where \(\alpha\), \(\beta\), and \(\gamma\) are nonnegative weighting parameters. The term \(\text{Acc}(\varphi(i))\) captures the classification accuracy for the chosen split, while the penalty terms reflect deficits in resources or time. Infeasible actions incur a direct penalty to discourage unworkable split assignments.

\subsubsection{RL Objective}
Each device \(d_i\) seeks a policy \(\pi_i\colon s_i^t \mapsto a_i^t\) that maximizes its discounted cumulative reward:
% \vspace{-0.15cm}
\begin{equation}
% \vspace{-0.15cm}
\max_{\pi_i} \quad \mathbb{E}_{\pi_i}\Bigl[\sum_{t=0}^{\infty}\,\delta^t\,r_i^t\Bigr],
\label{eq:rl_objective}
\end{equation}
where \(\delta \in [0,1]\) is discount factor. The Q-function \(Q_i^\pi(s, a)\) is approximated via a DQN, updated iteratively to converge to an optimal policy \(\pi_i^{*}\). This occurs for all devices in \(\mathcal{H}\).%, potentially with coordination by the server.

By penalizing or rewarding local splitting decisions, \textit{ReinDSplit} allocates deeper model segments to nodes with ample computational resources, while resource-constrained devices offload heavier workloads to a centralized server, eliminating conventional SL strategies' ``same-split-fits-all" limitation for heterogeneous scenarios with various applications such as the pest recognition system. Additionally, \textit{ReinDSplit} maintains raw image data onsite, preserving privacy and enabling large-scale deployment across geographically dispersed farms. 
We provide the pseudocode of \textit{ReinDSplit} in Algorithm~\ref{alg:ReinDSplit_insect_monitoring} as Appendix.

\section{Theoretical Analysis}\label{theoretical_ana}
This section develops a theoretical foundation for \emph{straggler mitigation} and \emph{convergence} of our \textit{ReinDSplit} framework.

\begin{definition}[\textbf{Straggler Effect in \textit{ReinDSplit}}]
\label{def:rl_straggler}
Let $\mathcal{H} = \{d_1, d_2, \dots, d_N\}$ be $N$ devices, and each $d_i$ selects a \textit{split index} $k \in \{1,\dots,K\}$ using its RL policy $\pi_i$. We define the \textit{straggler effect} as the probability that, in a training round $t$,
% \vspace{-0.2cm}
\begin{equation}
% \vspace{-0.25cm}
\exists\, d_i \;\text{such that}\; \Delta_i^k < 0
  \quad \text{and} \quad a_i^t = k,  
\end{equation}

where $\Delta_i^k$ is a local resource surplus for a device $d_i$ under split $k$. Equivalently, a \textit{straggler} arises if a device selects a split that exceeds its resource/time availability, delaying the global update or even incomplete local training.
\end{definition}

\begin{lemma}[Bound on Straggler Probability]
\label{lemma:straggler_probability}
Suppose each device $d_i$ executes a Q-learning over a finite state-action space $\mathcal{S}_i \times \{1,\dots, K\}$. Then the probability that $d_i$ selects an infeasible action $k \notin \mathcal{F}_i$ (i.e., $\Delta_i^k < 0$) $t\to 0$ as $t\to\infty$:
% \vspace{-0.15cm}
\begin{equation}
% \vspace{-0.15cm}
\lim_{t\to\infty}\; \Pr\bigl[a_i^t \notin \mathcal{F}_i\bigr] = 0  
\end{equation}

\end{lemma}
% \vspace{0.2cm}

\begin{theorem}[Diminishing Straggler Effect]
\label{theorem:straggler_diminish}
Let $\pi_i^{*}$ be the optimal policy for the device $d_i$ in our proposed \textit{ReinDSplit} framework. Suppose $\Pr[\text{Straggler at round }t]$ is the probability that, at time step $t$, at least one device $d_i \in \{d_1, \dots, d_N\}$ selects an action/split $k$ outside its feasibility set $\mathcal{F}_i$. Mathematically, 
\vspace{-0.35cm}
\begin{equation*}
    \vspace{-0.25cm}
  \Pr\bigl[\text{Straggler at round } t\bigr] 
  \;=\;
  \Pr\Bigl[\,
    \bigcup_{i=1}^{N}
    \{\,a_i^t \notin \mathcal{F}_i\}
  \Bigr].
\end{equation*}
Then, under Q-learning convergence: \vspace{-0.2cm}
\[
  \lim_{t\to\infty}\; \Pr\bigl[\text{Straggler at round } t\bigr] 
  \;=\; 0.
\]
\end{theorem}

\begin{definition}[\textbf{ReinDSplit Convergence}]
\label{def:global_rl_ReinDSplit_convergence}
\textit{Global convergence} is achieved when, across repeated training rounds, each device $d_i$ takes action $k \in \{1,\dots, K\}$ following its learned policy $\pi_i$, executes local forward/backward passes on $\mathcal{M}_C^k$, and transmits the resulting smashed gradients to the server for updates of $\mathcal{M}_S^k$. Convergence occurs if the local (client-side) and global (server-side) model parameters stabilize in expectation, ensuring no unbounded variance in performance metrics (accuracy) over time.
\end{definition}

\begin{lemma}[Stability of Local Updates]
\label{lemma:local_update_stability}
If each device $d_i$ consistently selects actions $k \in \mathcal{F}_i$ (its feasibility set), then local updates remain bounded. Formally, for any feasible split $k\in\mathcal{F}_i$, the gradients of the partial model $\mathcal{M}_C^k$ satisfy%\vspace{-0.1cm}
\[
  \|\nabla \mathcal{M}_C^k\| \;\le\; M_{\mathrm{grad}},
\]
where $M_{\mathrm{grad}}$ is a device-independent constant determined by batch size and network architecture factors.
\end{lemma}

\begin{theorem}[Global Convergence of \textit{ReinDSplit}]
\label{theorem:global_convergence_rl_ReinDSplit}
Suppose a finite MDP models each device $d_i$ with a corresponding Q-learning routine that converges to the optimal policy $\pi_i^{*}$; local gradients remain bounded as per Lemma~\ref{lemma:local_update_stability}; and a central server periodically aggregates partial updates from all devices. Then, our proposed \textit{ReinDSplit} algorithm converges in expectation to a stable partition of model parameters $\{\mathcal{M}_C^{k^*}, \mathcal{M}_S^{k^*}\}$ across devices, thereby maximizing total accuracy subject to the given resource/time constraints.
\end{theorem}

\begin{figure*}[!t]
	%	\begin{minipage}[]{1.0\linewidth}
			\centering	
            \includegraphics[width=1\linewidth]{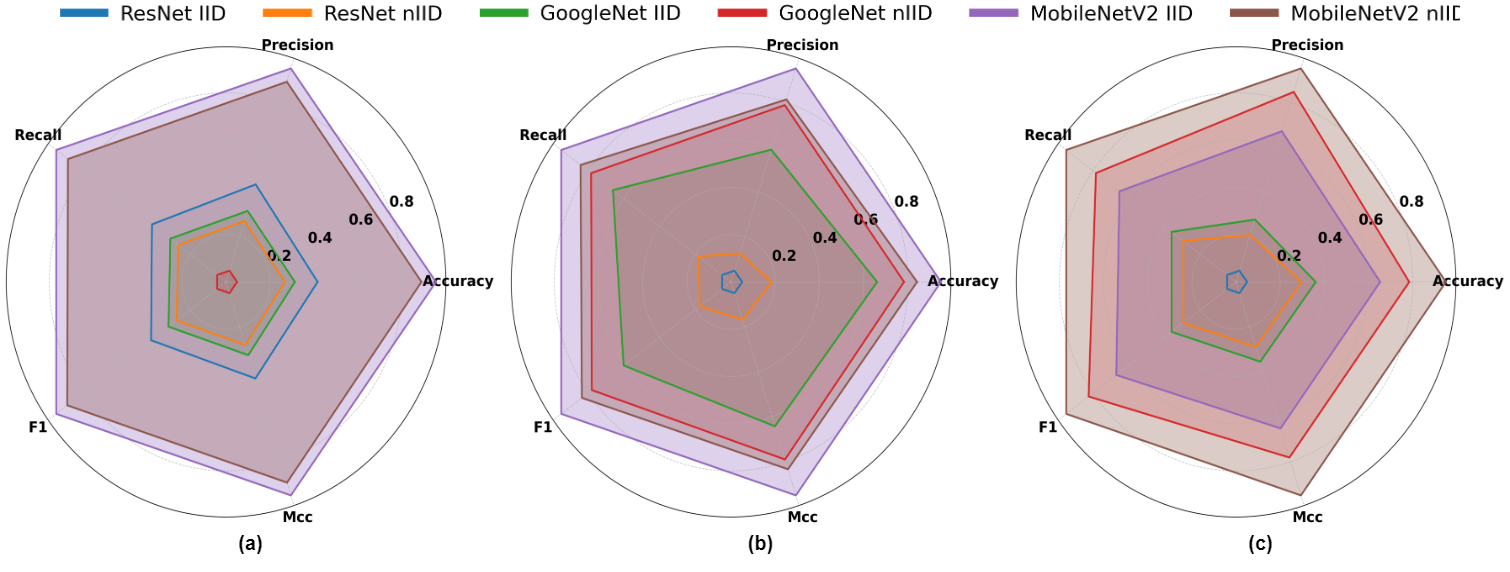}
			\vspace{-0.25in}
			\caption{Radar charts comparing normalized evaluation metrics for (a) EC, (b) FC, and (c) KAP datasets. Higher values closer to the outer edge indicate better performance, while the uniformity of the polygon shape reflects balanced model behavior across all metrics.}
			\vspace{-0.2in}
		\label{evaluation_metrics} 
	%	\end{minipage}
\end{figure*}

\section{Experimental Analysis}\label{experiments}%\vspace{-0.35cm}
To validate \textit{ReinDSplit}, we implemented all experiments in \textit{Python 3} using PyTorch, utilizing 2 Nvidia V100 GPU nodes, 8 CPU cores, and 40 GB of system memory. We performed our experiments on three DNN architectures: ResNet18 (RN), GoogleNet (GN), and MobileNetV2 (MN), and three pest datasets, namely Economic Crops (EC)~\cite{wu2019ip102}, Field Crops (FC)~\cite{wu2019ip102} (we adopted EC and FCs' classes as given in Table 1 of~\cite{wu2019ip102}), and Kaggle's Agriculture Pests (KAP)~\cite{kap} for clients = 10. For classes, please refer to Table~\ref{dataset_table} in the Appendix. We partitioned each dataset into train (75\%), validation (15\%), and test (10\%), and images are resized to \texttt{(224,224)}.% and normalized with mean~$[0.485,\,0.456,\,0.406]$ and standard deviation~$[0.229,\,0.224,\,0.225]$. 
To simulate the \textit{dynamic} split actions in vertical SL, we partition each model into \textbf{5 sub-models} for uniformly comparing computational loads between client and server when selecting different cut layers. For different split (or cut) layers of MobileNetV2, refer to Table~\ref{tab:MobileNetV2_splits} in the Appendix.\vspace{0.035cm} 

\noindent \textbf{Hyperparameter Tuning and Implementation:} We tuned hyperparameters with $20$ \textit{Optuna} trials: learning rate ($\text{lr = 1e-4, 1e-2}$), weight decay (= 1e-6, 1e-3), discount factor ($\gamma = [0.95, 0.99, 0.999]$), batch size (\{32, 64\}), and target network update frequency. Each trial runs for $50$ training episodes, with $75$ steps per episode. We mitigate oscillations in Q-learning with a replay buffer to batch updates of the Q-network, synchronizing a target network every $500$ or $1000$ step, depending on Optuna’s suggestion. We assigned a subset of the training dataset for each client via class-based shards to employ the non-IID distribution. We considered the cross-entropy loss function and the \texttt{AdamW} optimizer. We validated \textit{ReinDSplit}s' performance with accuracy, precision, recall, F1-score, and Matthews Correlation Coefficient (MCC).

\subsubsection{Heterogeneous Devices Simulation} To emulate farming environments, we simulate $N=5$ virtual clients, each with an assigned \textit{computational capacity} and \textit{time constraint} drawn from a continuous range $[0.5, 7.5]$). These states vary stochastically across training rounds to reflect dynamic resource availability, and each client can become \textit{unavailable} with 10\% probability. The agent’s \textit{action space} thus consists of $5$ possible split points, where choosing a higher split index implies more local layers. 

\begin{figure*}[!t]
	%	\begin{minipage}[]{1.0\linewidth}
			\centering	
            \includegraphics[width=1\linewidth]{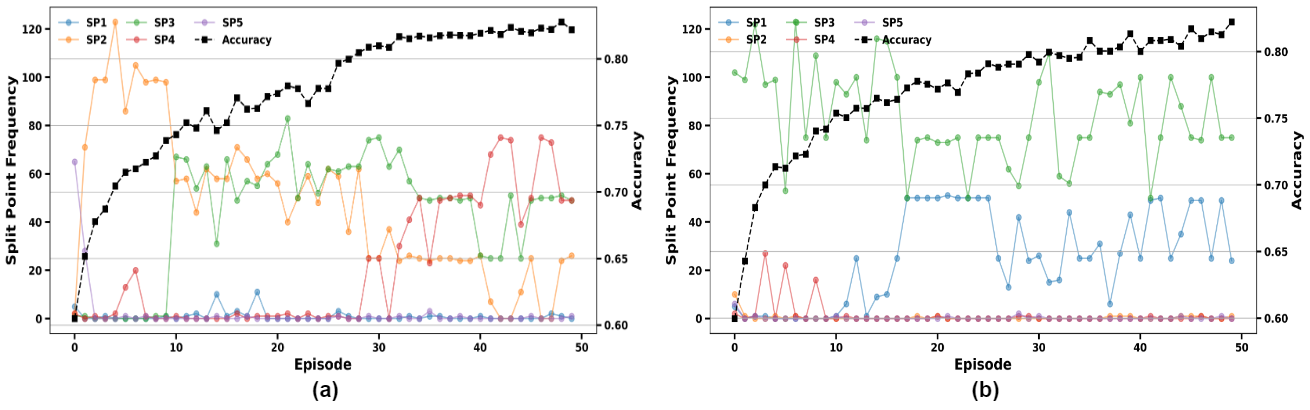}
			\vspace{-0.25in}
			\caption{Comparison of average split‐point frequency (left y‐axis) and validation accuracy (right y‐axis) over 50 episodes for the FC dataset under (a) IID and (b) non‐IID distributions. Each colored marker traces how often a given split point is selected, and the dashed line reflects the evolving mean accuracy.}
			\vspace{-0.25in}
		\label{split_freqencies_accuracy_trade-off} 
	%	\end{minipage}
\end{figure*}

\subsubsection{Dynamic Allocation} Our Q-learning framework maintains a state vector $\bigl[\text{compute\_capacity}, \text{time\_constraint}\bigr]$ for each client, and the agent selects one of the five split indices at each  \textit{round}. Our vanilla Q-network is a 2-layered, fully connected neural network with ReLU activation. Specifically, the first layer maps the 2-dimensional state vector  \(s_i^t \) to a 128-dimensional hidden representation, and the second layer outputs Q-values for each of the five possible splits. We deploy a $\epsilon$-greedy exploration strategy that decays over training episodes with the reward function defined in Eq.~\ref{eq:reward_function}.%\vspace{-0.15cm}

\subsection{Recognition Analysis} To visualize all evaluation metrics and DNN architectures, we aggregated the mean values for each metric in the IID and non-IID settings, as the radar presentations show in Fig.~\ref{evaluation_metrics}. Next, we applied the \textit{min-max} normalization approach, mapping all metrics to the $[0.01, 1]$ range to prevent zero values. This normalization highlights relative performance differences without skewing results toward high/low ranges.

\vspace{0.3em}
\noindent \textbf{EC:} MobileNetV2 achieved the highest maximum accuracy of $87.61\%$ (IID and $86.31\%$ (non-IID); thus, its observed range can be viewed as $79.19\%\pm8.42\%$. 
ResNet and GoogleNet followed closely with peak accuracies around $86.74\%$ (IID) and $86.37\%$ (IID). However, their non-IID performance dropped slightly to near $84.98\%$ and $86.31\%$, respectively. Across additional metrics (precision, recall, F1, MCC), all models consistently maintained values above $0.75$, signifying robust performance with resource-diverse devices. In comparison, traditional SL (with MobileNetV2) achieved maximum accuracies of 81\% (IID) and 79\% (non-IID) with precision, recall, and F1 scores around 0.76 (IID) and 0.74 (non-IID), while FL reached 90\% (IID) and 88\% (non-IID).

\vspace{0.3em}
\noindent \textbf{FC:} For this scenario, maximum accuracy spanned the interval $[79.24\%,\,82.78\%]$, with MobileNetV2 topping at $\approx82.78\%$ (IID) and $\approx82.24\%$ (non-IID). ResNet and GoogleNet followed suit, reaching $[79.24\%,\,81.53\%]$  range interval. Beyond accuracy, MobileNetV2’s precision, recall, and F1 lay in $[0.82,\,0.88]$ under both IID and non-IID conditions, edging out the competing models by a small but consistent margin. Hence, MobileNetV2 retained its advantage in recognizing pests despite varying data distributions. SL attained approximately 76.8\% (IID) and 76.2\% (non-IID), compared to 82.78\% (IID) and 82.24\% (non-IID) for \textit{ReinDSplit} and 85.0\% for FL.

\vspace{0.3em}
\noindent \textbf{KAP:} Here, MobileNetV2 again attained the top accuracy of roughly $94.31\%$ (IID), slightly dipping to $94.08\%$ (non-IID). Meanwhile, GoogleNet recorded a maximum near $93.62\%$, with a modest decline to about $93.20\%$ in non-IID. ResNet’s peak lay around $93.17\%$. When considering non-accuracy metrics, MobileNetV2’s F1 ranged from $0.90$ to $0.94$, and its MCC consistently exceeded $0.93$. Such stable, high metrics highlight MobileNetV2’s capability to handle large, heterogeneous datasets without significant performance degradation. Moreover, SL reported accuracy of 88.3\% (IID) and 88.1\% (non-IID), and for FL, 97.0\% (IID) and 96.5\% (non-IID), with F1 and MCC values following the same trend.
\vspace{-0.05cm}

\subsection{Impact of Split Points on Accuracy} We analyzed the trade-off of MobileNetV2 model split-point (SP) assignments to clients with validation accuracy for the FC dataset, as illustrated in Figure~\ref{split_freqencies_accuracy_trade-off}. In early episodes, SP2 often surges above 100 out of 120 possible selections, reflecting an initial strategy that balances computation and capacity. In the IID setting, SP2 dominates for the first 10 episodes, driving a 40-80\% accuracy range by episode 30 (ref. Figure~\ref{split_freqencies_accuracy_trade-off} (a)). However, SP3 progressively overtakes SP2 mid‐training, stabilizing near 70–80 selections while accuracy plateaus at around 78\%–80\%. In contrast, the non-IID setting triggers significant accuracy fluctuations—between 60\% and 78\%—as the model navigates heterogeneous distributions (Figure~\ref{split_freqencies_accuracy_trade-off} (b)). SP3 and SP4 each show 50‐point frequency swings between episodes 20 and 40, aligning with shifts in the loss landscape from uneven client splits. Meanwhile, SP1 and SP5 remain near zero frequency in both data distribution settings, indicating minimal performance gains despite occasional spikes (over 20) late in non-IID training. Finally, although both scenarios converge near 75\%–80\% accuracy, non-IID demonstrates greater volatility in split-point assignments, highlighting the need for adaptive partitioning when data distributions are non-uniform. 

\vspace{-0.15cm}
\subsection{Trade‐Offs Between Client Load, Reward, and Accuracy}
\begin{figure}[!h]
\vspace{-0.2cm}
		\begin{minipage}{\columnwidth}
		\centering			
  \includegraphics[width=1.0\linewidth]{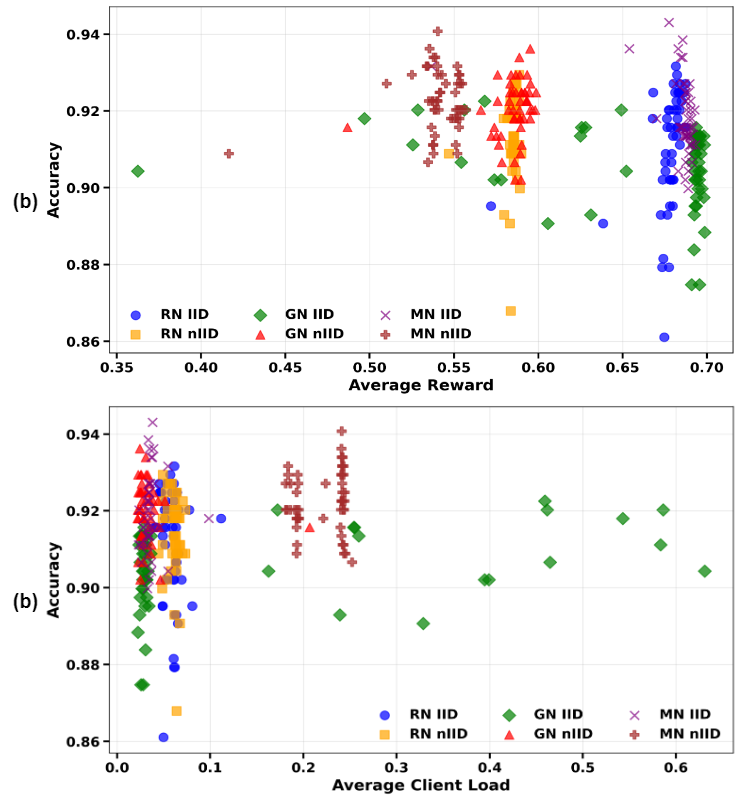}	
  \vspace{-0.6cm}
		\caption{\footnotesize{(a) Scatter plot showing average client load vs. model accuracy, and (b) scatter plot illustrating average reward vs. model accuracy, each for RN, GN, and MN under IID and non-IID settings.}}
  \vspace{-0.5cm}
		\label{reward_client_load_accuracy} 
		\end{minipage}
\end{figure}

In Fig.~\ref{reward_client_load_accuracy} (a) and (b), we examine how average reward (0.35–0.70) and average client load (0.0–0.60) correlate with classification accuracy (86\%–95\%) in \textit{ReinDSplit} for the KAP dataset. MN IID (purple ``\bm{$\times$}") occupies the upper‐right quadrant in both subplots, achieving 0.65–0.70 reward at 92\%–95\% accuracy and often surpassing 94\% accuracy at loads above 0.50, implying uniform data distributions leverage additional client computation effectively. In contrast, non-IID configurations—such as RN non-IID (orange squares)—cluster around 0.40–0.60 reward or below 0.3 load, with accuracies of 88\%–92\%, reflecting the constraints imposed by skewed data partitions. GN IID (green diamonds) strikes a balance in mid‐range reward (0.50–0.60) or load (0.2–0.5), frequently exceeding 90\% accuracy. Moreover, MN non-IID (pink plus signs) extends across moderate load levels (0.1–0.4) and reward values (0.45–0.65) while still reaching accuracy above 90\%, highlighting that architecture choice can partially offset heterogeneous data’s impact. We observed that allocating higher client loads boosts performance for IID scenarios. In contrast, non-IID settings require more adaptive strategies to maintain competitive accuracy. 

\section{Conclusion}\label{conclusion}%\vspace{-0.15cm}
This paper addressed the fundamental limitation of vertical split learning, where a uniform split point can bottleneck resource usage and degrade model convergence in heterogeneous device scenarios. By introducing \textit{ReinDSplit}, a Q-learning–based framework, we dynamically matched each client’s computational budget and time availability to an optimal partition of the neural network. Experimental results across four insect classification datasets showed that \textit{ReinDSplit} mitigates straggler issues and delivers accuracy in the $[71, 94]\%$. Our adaptive strategy preserved 82\%–94\% accuracy across diverse settings, even under non-IID conditions. These outcomes highlight \textit{ReinDSplit}’s potential for resource‐constrained agricultural applications, ensuring high recognition reliability without the pitfalls of a \textit{one‐size‐fits‐all} split. Moreover, by enabling on-the-fly split configuration, \textit{ReinDSplit} can adapt to the limited connectivity in rural farm environments, thus supporting timely pest surveillance and facilitating effective mitigation strategies in the smart connected farm network. In future work, we plan to incorporate bandwidth and energy constraints and explore compression techniques to optimize communication overhead, enabling faster training while maintaining competitive performance in large‐scale distributed environments. \vspace{0.1cm}

\noindent \textbf{Acknowledgement:} This work is supported by the National Science Foundation (NSF) Grants I-Corps (\#2331554), SIRAC (\#1952045), and PFI-RP (\#2431990), USA.

\bibliographystyle{IEEEbib}
\bibliography{my_bib}

\begin{thebibliography}{10}

\bibitem{I-Corps2023}
Vishesh~K. Tanwar,
\newblock ``I-{C}orps: Bugs-{B}-{G}one: Developing a robust and automated
  insect detection system for connecting farm communities,''
\newblock in {\em National Science Foundation (NSF) Award \# 2331554, USA},
  2023.

\bibitem{singh2024smart}
Asheesh~K Singh, Behzad~J Balabaygloo, et~al.,
\newblock ``Smart connected farms and networked farmers to improve crop
  production, sustainability and profitability,''
\newblock {\em Frontiers in Agronomy}, vol. 6, pp. 1410829, 2024.

\bibitem{antunes2022federated}
Rodolfo~Stoffel Antunes, Cristiano Andr{\'e}~da Costa, Arne K{\"u}derle,
  Imrana~Abdullahi Yari, and Bj{\"o}rn Eskofier,
\newblock ``Federated learning for healthcare: Systematic review and
  architecture proposal,''
\newblock {\em ACM Transactions on Intelligent Systems and Technology (TIST)},
  vol. 13, no. 4, pp. 1--23, 2022.

\bibitem{imteaj2021survey}
Ahmed Imteaj, Urmish Thakker, Shiqiang Wang, Jian Li, and M~Hadi Amini,
\newblock ``A survey on federated learning for resource-constrained iot
  devices,''
\newblock {\em Internet of Things Journal}, vol. 9, no. 1, pp. 1--24, 2021.

\bibitem{mothukuri2021survey}
Viraaji Mothukuri, Reza~M Parizi, Seyedamin Pouriyeh, Yan Huang, Ali
  Dehghantanha, and Gautam Srivastava,
\newblock ``A survey on security and privacy of federated learning,''
\newblock {\em Future Generation Computer Systems}, vol. 115, pp. 619--640,
  2021.

\bibitem{vepakomma2018split}
Praneeth Vepakomma, Otkrist Gupta, Tristan Swedish, and Ramesh Raskar,
\newblock ``Split learning for health: Distributed deep learning without
  sharing raw patient data,''
\newblock {\em arXiv preprint arXiv:1812.00564}, 2018.

\bibitem{bakhtiarnia2023dynamic}
Arian Bakhtiarnia, Nemanja Milo{\v{s}}evi{\'c}, Qi~Zhang, Dragana Bajovi{\'c},
  and Alexandros Iosifidis,
\newblock ``Dynamic split computing for efficient deep edge intelligence,''
\newblock in {\em IEEE international conference on acoustics, speech and signal
  processing}. IEEE, 2023, pp. 1--5.

\bibitem{cao2024learning}
Yang Cao, Shao-Yu Lien, Cheng-Hao Yeh, Der-Jiunn Deng, Ying-Chang Liang, and
  Dusit Niyato,
\newblock ``Learning-based multi-tier split computing for efficient convergence
  of communication and computation,''
\newblock {\em IEEE Internet of Things Journal}, 2024.

\bibitem{lima2020automatic}
Matheus Lima, Maria Leandro, Constantino Valero, Luis Coronel, and Clara Bazzo,
\newblock ``Automatic detection and monitoring of insect pests—a review,''
\newblock {\em Agriculture}, vol. 10, no. 5, pp. 161, 2020.

\bibitem{butera2021precise}
Luca Butera, Alberto Ferrante, Mauro Jermini, Mauro Prevostini, and Cesare
  Alippi,
\newblock ``Precise agriculture: effective deep learning strategies to detect
  pest insects,''
\newblock {\em IEEE/CAA Journal of Automatica Sinica}, vol. 9, no. 2, pp.
  246--258, 2021.

\bibitem{sishodia2020applications}
Rajendra~P Sishodia, Ram~L Ray, and Sudhir~K Singh,
\newblock ``Applications of remote sensing in precision agriculture: A
  review,''
\newblock {\em Remote sensing}, vol. 12, no. 19, pp. 3136, 2020.

\bibitem{kirkeby2021advances}
Carsten Kirkeby, Klas Rydhmer, Samantha~M Cook, Alfred Strand, Martin~T
  Torrance, Jennifer~L Swain, Jord Prangsma, Andreas Johnen, Mikkel Jensen,
  Mikkel Brydegaard, et~al.,
\newblock ``Advances in automatic identification of flying insects using
  optical sensors and machine learning,''
\newblock {\em Scientific reports}, vol. 11, no. 1, pp. 1555, 2021.

\bibitem{iost2020drones}
Fernando~H Iost~Filho, Wieke~B Heldens, Zhaodan Kong, and Elvira~S de~Lange,
\newblock ``Drones: innovative technology for use in precision pest
  management,''
\newblock {\em Journal of economic entomology}, vol. 113, no. 1, pp. 1--25,
  2020.

\bibitem{velusamy2021unmanned}
Parthasarathy Velusamy, Santhosh Rajendran, Rakesh~Kumar Mahendran, Salman
  Naseer, Muhammad Shafiq, and Jin-Ghoo Choi,
\newblock ``Unmanned aerial vehicles (uav) in precision agriculture:
  Applications and challenges,''
\newblock {\em Energies}, vol. 15, no. 1, pp. 217, 2021.

\bibitem{albanese2021automated}
Andrea Albanese, Matteo Nardello, and Davide Brunelli,
\newblock ``Automated pest detection with dnn on the edge for precision
  agriculture,''
\newblock {\em IEEE Journal on Emerging and Selected Topics in Circuits and
  Systems}, vol. 11, no. 3, pp. 458--467, 2021.

\bibitem{chen2020aiot}
Ching-Ju Chen, Ya-Yu Huang, Yuan-Shuo Li, Chuan-Yu Chang, and Yueh-Min Huang,
\newblock ``An aiot based smart agricultural system for pests detection,''
\newblock {\em IEEE access}, vol. 8, pp. 180750--180761, 2020.

\bibitem{anand2021agrisegnet}
Tanmay Anand, Soumendu Sinha, Murari Mandal, Vinay Chamola, and Fei~Richard Yu,
\newblock ``Agrisegnet: Deep aerial semantic segmentation framework for
  iot-assisted precision agriculture,''
\newblock {\em IEEE Sensors Journal}, vol. 21, no. 16, pp. 17581--17590, 2021.

\bibitem{pham2023binarizing}
Ngoc~Duy Pham, Alsharif Abuadbba, Yansong Gao, Khoa~Tran Phan, and Naveen
  Chilamkurti,
\newblock ``Binarizing split learning for data privacy enhancement and
  computation reduction,''
\newblock {\em IEEE Transactions on Information Forensics and Security}, vol.
  18, pp. 3088--3100, 2023.

\bibitem{zheng2024ppsfl}
Jiali Zheng, Yixin Chen, and Qijia Lai,
\newblock ``Ppsfl: Privacy-preserving split federated learning for
  heterogeneous data in edge-based internet of things,''
\newblock {\em Future Generation Computer Systems}, vol. 156, pp. 231--241,
  2024.

\bibitem{zhang2023privacy}
Zongshun Zhang, Andrea Pinto, Valeria Turina, Flavio Esposito, and Ibrahim
  Matta,
\newblock ``Privacy and efficiency of communications in federated split
  learning,''
\newblock {\em IEEE Transactions on Big Data}, vol. 9, no. 5, pp. 1380--1391,
  2023.

\bibitem{lin2024efficient}
Zheng Lin, Guangyu Zhu, Yiqin Deng, Xianhao Chen, Yue Gao, Kaibin Huang, and
  Yuguang Fang,
\newblock ``Efficient parallel split learning over resource-constrained
  wireless edge networks,''
\newblock {\em IEEE Transactions on Mobile Computing}, 2024.

\bibitem{shen2023ringsfl}
Jinglong Shen, Nan Cheng, Xiucheng Wang, Feng Lyu, Wenchao Xu, Zhi Liu, Khalid
  Aldubaikhy, and Xuemin Shen,
\newblock ``Ringsfl: An adaptive split federated learning towards taming client
  heterogeneity,''
\newblock {\em IEEE Transactions on Mobile Computing}, 2023.

\bibitem{tianqing2021resource}
Zhu Tianqing, Wei Zhou, Dayong Ye, Zishuo Cheng, and Jin Li,
\newblock ``Resource allocation in iot edge computing via concurrent federated
  reinforcement learning,''
\newblock {\em IEEE Internet of Things Journal}, vol. 9, no. 2, pp. 1414--1426,
  2021.

\bibitem{xiong2020resource}
Xiong Xiong, Kan Zheng, Lei Lei, and Lu~Hou,
\newblock ``Resource allocation based on deep reinforcement learning in iot
  edge computing,''
\newblock {\em IEEE Journal on Selected Areas in Communications}, vol. 38, no.
  6, pp. 1133--1146, 2020.

\bibitem{liu2020resource}
Xiaolan Liu, Jiadong Yu, Jian Wang, and Yue Gao,
\newblock ``Resource allocation with edge computing in iot networks via machine
  learning,''
\newblock {\em IEEE Internet of Things Journal}, vol. 7, no. 4, pp. 3415--3426,
  2020.

\bibitem{yan2020offloading}
Jia Yan, Suzhi Bi, and Ying Jun~Angela Zhang,
\newblock ``Offloading and resource allocation with general task graph in
  mobile edge computing: A deep reinforcement learning approach,''
\newblock {\em IEEE Transactions on Wireless Communications}, vol. 19, no. 8,
  pp. 5404--5419, 2020.

\bibitem{alqerm2021deepedge}
Ismail AlQerm and Jianli Pan,
\newblock ``Deepedge: A new qoe-based resource allocation framework using deep
  reinforcement learning for future heterogeneous edge-iot applications,''
\newblock {\em IEEE Transactions on Network and Service Management}, vol. 18,
  no. 4, pp. 3942--3954, 2021.

\bibitem{liu2020multi}
Xiaolan Liu, Jiadong Yu, Zhiyong Feng, and Yue Gao,
\newblock ``Multi-agent reinforcement learning for resource allocation in iot
  networks with edge computing,''
\newblock {\em China Communications}, vol. 17, no. 9, pp. 220--236, 2020.

\bibitem{wu2019ip102}
Xiaoping Wu, Chi Zhan, Yu-Kun Lai, Ming-Ming Cheng, and Jufeng Yang,
\newblock ``Ip102: A large-scale benchmark dataset for insect pest
  recognition,''
\newblock in {\em IEEE/CVF conference on computer vision and pattern
  recognition}, 2019, pp. 8787--8796.

\bibitem{kap}
Kaggle,
\newblock ``{Agricultural Pests Image Dataset},''
  \url{https://www.kaggle.com/datasets/vencerlanz09/agricultural-pests-image-dataset},
\newblock [Online; accessed January 02, 2025].

\end{thebibliography}

\newpage

\section*{Appendix}

\section*{Our Proposed Framework}
\begin{algorithm}[!h]
\setlength{\floatsep}{2pt}      
\setlength{\textfloatsep}{2pt}  
\setlength{\intextsep}{2pt}     
\caption{\textit{ReinDSplit} Algorithm}
\label{alg:ReinDSplit_insect_monitoring}
\begin{algorithmic}[1]
\REQUIRE $\mathcal{H}=\{d_1,\dots,d_N\}$, total rounds $T$, Q-net params $\{\theta_i\}$, $\epsilon$, $\gamma$, $\mathcal{A}_i=\{1,\dots,K\}$.
\ENSURE $\{\theta_i^*\}$ for split decisions.
\vspace{2pt}
\FOR{$t=1 \rightarrow T$}
    \FOR{each $d_i \in \{1,\dots,N\}$ \textbf{in parallel}}
        \STATE Observe $s_i^t \!=\!(R_i^t,\,T_i^t,\,\mathcal{P}_i^t)$
        \STATE Select $a_i^t$:\vspace{-0.4cm}
        \[
          a_i^t = \begin{cases}
          \text{random in } \mathcal{A}_i, & \text{w.p.\ } \epsilon,\\
          \arg\max\limits_{a} Q_i(s_i^t,a;\theta_i), & \text{otherwise}
          \end{cases}
        \]%\vspace{-0.3cm}
        \IF{$R_i^t \!\ge\! R_{\mathrm{req}}(\mathcal{M}_C^{a_i^t}) \!\land\! T_i^t \!\ge\! T_{\mathrm{req}}(\mathcal{M}_C^{a_i^t})$}
            \STATE \textit{Local training} on $\mathcal{M}_C^{a_i^t}$; send smashed data
            \STATE \textit{Server update} on $\mathcal{M}_S^{a_i^t}$; return gradients
            \STATE \textit{Local backprop}; compute $r_i^t$
        \ELSE
            \STATE $r_i^t \!\leftarrow\! -\gamma\,(\text{penalty})$
        \ENDIF
        \STATE Update $Q_i$ with $(s_i^t,\,a_i^t,\,r_i^t,\,s_i^{t+1})$: %\vspace{-0.2cm}
        \[
          L(\theta_i)\!=\!\bigl(r_i^t + \gamma\,\max_{a'}Q_i(s_i^{t+1},a';\theta_i^{-}) - Q_i(s_i^t,a_i^t;\theta_i)\bigr)^2
        \]\vspace{-0.25cm}
        \STATE Periodically set $\theta_i^{-} \leftarrow \theta_i$
    \ENDFOR
    \STATE $\epsilon \leftarrow \text{decay}(\epsilon)$
\ENDFOR
\RETURN $\{\theta_i\}$
\end{algorithmic}
% \vspace{-0.15cm}
\end{algorithm}

\section*{Theoretical Analysis}\label{theoretical_ana}

\begin{proof}[Proof of Lemma 1] Given $d_i$ interacts with a finite state-action space \(\mathcal{S}_i \times \{1,\dots,K\}\) and each action \(k \in \{1,\dots,K\}\) corresponds to a specific client-side split \(\mathcal{M}_C^k\) and server-side \(\mathcal{M}_S^k\). With the reward function in Eq.~\eqref{eq:reward_function}, selecting an infeasible action \(k \notin \mathcal{F}_i\) results in a substantial negative reward due to resource or time constraints being violated (\(\Delta_i^k < 0\)). Conversely, feasible action \(k \in \mathcal{F}_I \) yields higher or non-negative rewards, reflecting successful model training without resource violations.

Given that the state and action spaces are finite, Q-learning convergence results (ref. \ Section~\ref{prelim:rl}) ensure that the Q-values \(Q_i(s, a)\) will converge to the optimal Q-function \(Q_i^{*}(s, a)\) provided that each state-action pair is visited sufficiently often and the learning rate satisfies the usual diminishing conditions. Constantly associated with negative rewards, infeasible actions will have their Q-values suppressed over time relative to feasible actions.

As a result, the optimal policy \(\pi_i^{*}\), derived from the converged Q-function, will preferentially select actions from the feasibility set \(\mathcal{F}_i\). Formally, for any state \(s \in \mathcal{S}_i\),
% \vspace{-0.2cm}
\begin{equation}
% \vspace{-0.2cm}
\pi_i^{*}(s) = \arg\max_{k \in \mathcal{F}_i} Q_i^{*}(s, k),    
\end{equation}
ensuring that infeasible actions are never chosen in the limit. Consequently, the probability that \(d_i\) selects an infeasible action \(k \notin \mathcal{F}_i\) approaches zero as the number of training rounds \(t\) increases: $\lim_{t \to \infty} \Pr\bigl[a_i^t \notin \mathcal{F}_i\bigr] = 0.$
\end{proof}

\begin{proof}[Proof of Theorem 1]
For each device $d_i$, Lemma~\ref{lemma:straggler_probability} establishes that %\vspace{-0.1cm}
\[
  \lim_{t\to\infty}\,\Pr\bigl[a_i^t \notin \mathcal{F}_i\bigr] 
  \;=\;
  0.
\]
In other words, the probability of selecting an infeasible action for any individual device $d_i$ vanishes as $t$ grows. By a standard union bound, we have %\vspace{-0.2cm}
\[
  \Pr\Bigl[
    \bigcup_{i=1}^N \{a_i^t \notin \mathcal{F}_i\}
  \Bigr]
  \;\le\;
  \sum_{i=1}^{N}
    \Pr\bigl[a_i^t \notin \mathcal{F}_i\bigr].
\]
Since each term $\Pr[a_i^t \notin \mathcal{F}_i]$ converges to zero by Lemma~\ref{lemma:straggler_probability}, the sum on the right also converges to zero. Consequently,\vspace{-0.2cm}
\[
  \lim_{t\to\infty}\,\Pr\Bigl[
    \bigcup_{i=1}^N \{a_i^t \notin \mathcal{F}_i\}
  \Bigr]
  \;=\;
  0.
\]
Hence, the probability of encountering \textit{any} straggler in a given round diminishes to zero in the limit. 
\end{proof}

\begin{proof}[Proof of Lemma 2]
Since $d_i$ selects only $k\in \mathcal{F}_i$ and by definition of feasibility, $d_i$ can complete the local forward and backward passes without interruptions or resource/time deficits. In this setting, the assumptions on stochastic gradient descent (see Section~\ref{prelim:rl}) imply that the norm of each gradient update is bounded when the input batch size and network layers are finite and \textit{Lipschitz} continuous.  

Mathematically, let $\mathcal{L}\bigl(\mathcal{M}_C^k; \mathbf{x}_i, \mathbf{y}_i\bigr)$ be the local loss function for device $d_i$, based on training batch-size $\{(\mathbf{x}_i, \mathbf{y}_i)\}$. Under mild smoothness conditions on $\mathcal{L}$, it follows that%\vspace{-0.1cm}
\[
  \bigl\|\nabla \mathcal{M}_C^k\bigr\|
  \;=\;
  \Bigl\|
    \frac{\partial \mathcal{L}}{\partial \mathcal{M}_C^k}
    \Bigl(\mathbf{x}_i, \mathbf{y}_i\Bigr)
  \Bigr\|
  \;\le\;
  M_{\mathrm{grad}},
\]
where $M_{\mathrm{grad}}$ is constant depending on the maximum batch size, the network’s layer structure, and \textit{Lipschitz} coefficients of the activation functions. Because the device remains within its feasible resource/time range ($k \in \mathcal{F}_i$), it completes backpropagation successfully, preserving this bound each round.

Thus, when each $d_i$ executes feasibility constraints, the partial model updates $\nabla \mathcal{M}_C^k$ cannot grow unbounded. This proves stable local learning and supports global RL-\textit{ReinDSplit} convergence, as no erratic or divergent local updates pollute the server’s aggregated model.
\end{proof}

\begin{proof}[Proof of Theorem 2]
Given each device $d_i$ learns via Q-learning over a finite state-action space, selecting a split $k\in\{1,\dots, K\}$ to execute locally. By convergence analysis for Q-learning in finite MDPs, each device’s action-value function $Q_i$ converges to an optimal $Q_i^{*}$. Lemma~\ref{lemma:straggler_probability} proved that it yields an optimal policy $\pi_i^{*}$ that avoids infeasible splits over continuous learning.

Next, by Lemma~\ref{lemma:local_update_stability}, whenever a device chooses a feasible split $k \in \mathcal{F}_i$, its local gradient norm remains bounded by $M_{\mathrm{grad}}$. Consequently, in each training round, the server receives non-explosive partial gradients, aggregates them for the server-side model $\mathcal{M}_S^k$, and periodically synchronizes its updated parameters with all participating devices.

Over training episodes, no device selects an infeasible split with high probability, as proved in Theorem~\ref{theorem:straggler_diminish}, thereby preventing any ``straggling" and ensuring stable local parameter updates. Since each device’s local Q-learning policy converges to maximizing classification accuracy (with resource/time limits), the aggregated global model $\{\mathcal{M}_C^{k}, \mathcal{M}_S^{k}\}$ converges to a partition that yields consistently high performance across all devices. In expectation, no unbounded fluctuations remain in either local or server-side parameters, ensuring convergence in expectation to a stable global partition $\{\mathcal{M}_C^{k^*}, \mathcal{M}_S^{k^*}\}$. Therefore, with the reward structure and Q-value maximization, this partition balances accuracy with resource/time constraints for each device.
\end{proof}

\section*{Experimental Analysis}\label{experiments}%\vspace{-0.35cm}

\begin{table}[!h]
% \vspace{-0.3cm}
\centering
\caption{Overview of Pest Datasets}
% \vspace{-0.15cm}
\label{dataset_table}
\begin{tabular}{|c|c|l|}
\hline
\multicolumn{1}{|c|}{\textbf{Dataset}} & \textbf{\# Classes} & \multicolumn{1}{c|}{\textbf{Class Labels}} \\ \hline
EC & 3 & Vitis, Citrus, Mango                       \\ \hline
FC & 5 & Rice, Corn, Wheat, Beet, Alfalfa           \\ \hline
\begin{tabular}[c]{@{}l@{}}KAP\end{tabular} & 12 & \begin{tabular}[c]{@{}l@{}}Ants, Bees, Beetles, Caterpillars, \\ Moth, Earthworms, Earwigs, \\ Grasshoppers, Slugs, Snails,\\ Wasps, Weevils.\end{tabular} \\ \hline
\end{tabular}
% \vspace{-0.25cm}
\end{table}

\begin{table}[!h]
% \vspace{-0.15cm}
\centering
\caption{MobileNetV2 Partitioned into 5 Splits}
\label{tab:MobileNetV2_splits}
% \vspace{-0.15cm}
\begin{tabular}{|c|p{0.7\columnwidth}|}
\hline
\textbf{Split Index} & \textbf{Client-Side Layers} \\
\hline
1 & \texttt{features[0]} \newline (\textit{Initial Conv, BatchNorm, ReLU}) \\ \hline
2 & \texttt{features[0..3]} \newline (\textit{Initial + 3 Residual Blocks}) \\ \hline
3 & \texttt{features[0..7]} \newline (\textit{Initial + 7 Residual Blocks}) \\ \hline
4 & \texttt{features[0..13]} \newline (\textit{Initial + 13 Residual Blocks}) \\ \hline
5 & \texttt{features[0..18]} + Global Avg Pooling \\ \hline
\end{tabular}
% \vspace{-0.7cm}
\end{table}

\end{document}